# Chi-square Tests Driven Method for Learning the Structure of Factored MDPs

**Thomas Degris**
Thomas.Degris@lip6.fr

**Olivier Sigaud**
Olivier.Sigaud@lip6.fr

**Pierre-Henri Wuillemin**
Pierre-Henri.Wuillemin@lip6.fr

University Pierre et Marie Curie - Paris 6, LIP6
4 Place Jussieu
75252 Paris Cedex 05, France

## Abstract

SDYNA is a general framework designed to address large stochastic reinforcement learning (RL) problems. Unlike previous model-based methods in Factored MDPs (FMDPs), it incrementally learns the structure of a RL problem using supervised learning techniques. SPITI is an instantiation of SDYNA that uses decision trees as factored representations. First, we show that, in structured RL problems, SPITI learns the structure of FMDPs using Chi-Square tests and performs better than classical tabular model-based methods. Second, we study the generalization property of SPITI using a Chi-Square based measure of the accuracy of the model built by SPITI.

## 1 INTRODUCTION

A Markov Decision Process (MDP) is a mathematical framework providing a semantic model for planning under uncertainty. Solution methods to solve MDPs based on Dynamic Programming (DP) are known to work well on small problems but rely on explicit state space enumeration. Factored MDPs (FMDPs) are an alternative framework to compactly represent large and structured MDPs (Boutilier et al., 1995). In this framework, *Dynamic Bayesian Networks* (DBNs) (Dean & Kanazawa, 1989) are used to represent compactly the transition and the reward functions by exploiting the dependencies between variables composing a state.

Solution methods adapted from classical techniques such as DP or Linear Programming (LP) have been proposed (Hoey et al., 1999; Boutilier et al., 2000; Guestrin et al., 2003) and successfully tested on large stochastic planning problems. Moreover, model-based learning methods (Guestrin et al., 2002) have been proposed to learn the parameters of the model within the FMDP framework, assuming that the structure of the problem is known.

When the transition and reward functions are unknown, learning in FMDPs becomes a problem of learning the structure of DBNs from experience only. Chickering et al. (1997) and Friedman and Goldszmidt (1998) investigate techniques for learning Bayesian networks with local structure. From a given set of observations, these techniques explicitly learn the global structure of the network and the local structures quantifying this network.

In this paper, we describe *Structured* DYNA (SDYNA), a general framework merging supervised learning algorithms with planning methods within the FMDP framework. SDYNA does not require to explicitly build the global structure of the DBNs. More precisely, we focus on SPITI, an instantiation of SDYNA, that uses decision trees as factored representations. SPITI directly learns local structures of DBNs based on an incremental decision tree induction algorithm, namely ITI (*Incremental Tree Inducer* (Utgoff et al., 1997)). These representations are simultaneously exploited by a modified incremental version of the *Structured Value Iteration* (SVI) algorithm (Boutilier et al., 2000) computing a policy executed by the agent.

In the context of incremental decision tree induction, we use $\chi^2$ as a test of independence between two probability distributions (Quinlan, 1986) to build a model of the transition and reward functions. The $\chi^2$ threshold used to split two distributions directly impact the size of the model learned and may be critical in finding a good policy. First, we show that, setting the $\chi^2$ threshold to a high value makes SPITI able to build a compact model without impacting the quality of its policy. Second, we show that, while keeping its model compact, SPITI exploits the generalization property in its model learning method to perform better than a stochastic version of DYNA-Q (Sutton, 1990), a tabular model-based RL method. Finally, we introduce a new measure of the accuracy of the transition model based on $\chi^2$ to study the generalization property of SPITI. We show that the accuracy of the model learned by SPITI decreases linearly when the size of the problem grows exponentially.

The remainder of this paper is organized as follows: in Section 2, we introduce FMDPs. In Section 3, we describe the SDYNA framework. In Section 4, we describe SPITI and explain how we exploit $\chi^2$ in model learning and evaluation. Section 5 describes empirical results with SPITI. We discuss these results in Section 6. We conclude and describe some future work within the SDYNA framework in Section 7.

# 2 BACKGROUND

A MDP is defined by a tuple $\langle S, A, R, P \rangle$. $S$ is a finite set of states, $A$ is a finite set of actions, $R$ is the immediate *reward function* with $R : S \times A \rightarrow \mathbb{R}$ and $P$ is the *Markovian transition function* $P(s'|s,a)$ with $P : S \times A \times S \rightarrow [0,1]$. A *stationary policy* $\pi$ is a mapping $S \rightarrow A$ with $\pi(s)$ defining the action to be taken in state $s$. Considering an infinite horizon, we evaluate a policy $\pi$ in state $s$ with the *value function* $V_\pi(s)$ defined using the discounted reward criterion: $V_\pi(s) = E_\pi[\sum_{t=0}^{\infty} \gamma^t \cdot r_t | s_0 = s]$, with $0 \leq \gamma < 1$ the discount factor and $r_t$ the reward obtained at time $t$. The action-value function $Q_\pi(s,a)$ is defined as:

$$Q_\pi(s,a) = \sum_{s' \in S} P(s'|s,a)(R(s',a) + \gamma V_\pi(s')) \qquad (1)$$

A policy $\pi$ is optimal if for all $s \in S$ and all policies $\pi'$, $V_\pi > V_{\pi'}$. The value function of any optimal policy is called the *optimal value function* and is noted $V^*$.

The factored MDP framework (Boutilier et al., 1995) is a representation language exploiting the structure of the problem to represent compactly large MDPs with factored representations. In a FMDP, states are composed of a set of random variables $\{X_1, \ldots, X_n\}$ with each $X_i$ taking its value in a finite domain $Dom(X_i)$. A state is defined by a vector of values $s = (x_1, \ldots, x_n)$ with for all $i$: $x_i \in Dom(X_i)$. The state transition model $T_a$ of an action $a$ is defined by a transition graph $G_a$ represented as a DBN (Dean & Kanazawa, 1989). $G_a$ is a two-layer directed acyclic graph whose nodes are $\{X_1, \ldots, X_n, X'_1, \ldots, X'_n\}$ with $X_i$ a variable at time $t$ and $X'_i$ the same variable at time $t+1$. The parents of $X'_i$ are noted $Parents_a(X'_i)$. We assume that $Parents_a(X'_i) \subseteq X$ (i.e. there are no *synchronic* arcs, that is arcs from $X'_i$ to $X'_j$). A graph $G_a$ is quantified by a *Conditional Probability Distribution* $\mathrm{CPD}^a_{X_i}(X'_i|Parents_a(X'_i))$ associated to each node $X'_i \in G_a$. The transition model $T$ of the FMDP is then defined by a separate DBN model $T_a = \langle G_a, \{\mathrm{CPD}^a_{X_1}, \ldots, \mathrm{CPD}^a_{X_n}\}\rangle$ for each action $a$.

# 3 SDYNA

Similarly to the DYNA architecture (Sutton, 1990), SDYNA proposes to integrate planning, acting and learning to solve by trial-and-error stochastic RL problem with unknown transition and reward functions. However, SDYNA uses FMDPs as a representation language to be able to address large RL problems. An overview of SDYNA is given in Figure 1.

Input: *Acting*, *Learn*, *Plan*, Fact
Output: $\emptyset$

1. Initialization
2. At each time step $t$, do:
   (a) $s \leftarrow$ current (non-terminal) state
   (b) $a \leftarrow Acting(s, \{\mathrm{Fact}(Q_{t-1}(s,a)), a \in A\})$
   (c) Execute $a$; observe $s'$ and $r$
   (d) $\mathrm{Fact}(M_t) \leftarrow Learn(\mathrm{Fact}(M_{t-1}), s, a, s', r)$
   (e) $\{\mathrm{Fact}(V_t), \{\mathrm{Fact}(Q_t(s,a)), a \in A\}\} \leftarrow$ $Plan(\mathrm{Fact}(M_t), \mathrm{Fact}(V_{t-1}))$

   with $M_t$ the model of the transition and reward functions at time $t$.

Figure 1: The SDYNA algorithm

Neither the FMDP framework nor SDYNA specify which factored representations should be used. Factored representations, noted $\mathrm{Fact}(F)$ in SDYNA, can exploit certain regularities in the represented function $F$. These representations include rules, decision trees or algebraic decision diagrams. SDYNA is decomposed in three phases: acting (steps 2.a, 2.b and 2.c), learning (steps 2.d) and planning (steps 2.e). The next section details these phases in the context of an instantiation of SDYNA named SPITI.

# 4 SPITI

SPITI uses decision trees as factored representations (noted $\mathrm{Tree}(F)$). Section 4.1 and Section 4.2 describe, respectively, the acting and planning phases and then the learning phase in SPITI.

## 4.1 ACTING AND PLANNING

Acting in SPITI is similar to acting in other RL algorithms. The planning phase builds the set $S_Q$ of action-value functions $\mathrm{Tree}(Q_{t-1}(s,a))$ representing the expected discounted reward for taking action $a$ in state $s$ and then following a greedy policy. Thus, the agent can behave greedily by executing the best action in all states. SPITI uses the $\epsilon$-*greedy* exploration policy which executes the best action most of the time, and, with a small probability $\epsilon$, selects uniformly at random an action, independently of $S_Q$.

Planning has been implemented using an incremental version of the SVI algorithm (Boutilier et al., 2000). SVI is adapted from the Value Iteration algorithm but using decision trees as factored representations instead of tabular representations. SPITI uses an incremental version of SVI rather than the original SVI for two reasons. First, SVI returns a greedy policy, which may not be adequate for exploration policies other than $\epsilon$-*greedy*. Second, SVI computes an evaluation of the value function until convergence despite an incomplete and incrementally updated model of the transition and reward functions. The modified version of SVI used in SPITI is described in Figure 2.

---

Input: $\text{Tree}(M), \text{Tree}(T_t), \text{Tree}(V_{t-1})$
Output: $\text{Tree}(V_t), \{\text{Tree}(Q_t(s,a)), a \in A\}$

1. $S_Q = \{\text{Tree}(Q_t(s,a)), a \in A\}$ with: $\text{Tree}(Q_t(s,a)) \leftarrow Regress(\text{Tree}(M), \text{Tree}(V_{t-1}), a)$
2. $\text{Tree}(V_t) \leftarrow Merge(S_Q)$ (using maximization over the value as combination function).
3. Return $\{\text{Tree}(V_t), S_Q\}$

---

Figure 2: SPITI (1): the *Plan* algorithm based on SVI.

At each time step, the *Plan* algorithm in SPITI updates set $S_Q$ by producing the action-value function $\text{Tree}(Q_t(s,a))$ with respect to the value function $\text{Tree}(V_{t-1})$ using the *Regress* operator (step 1) defined in Boutilier et al. (2000). Then, action-value functions $\text{Tree}(Q_t(s,a))$ are merged using maximization as combination function to compute the value function $\text{Tree}(V_t)$ associated with a greedy policy using the $Merge(\{T_1, \ldots, T_n\})$ operator. This operator is used to produce a single tree containing all the partitions occurring in all trees $T_1, \ldots, T_n$ to be merged, and whose leaves are labeled using a *combination function* of the labels of the corresponding leaves in the original trees. $\text{Tree}(V_t)$ is reused at time $t+1$ to update the set $S_Q$ of action-value functions $\text{Tree}(Q_{t+1}(s,a))$. We refer to Boutilier et al. (2000) for a detailed description of the *Merge* and *Regress* operators.

### 4.2 LEARNING THE STRUCTURE

Trials of the agent compose a stream of examples that must be learned incrementally. In SPITI, we use incremental decision tree induction algorithms (Utgoff, 1986), noted *LearnTree*. From a stream of examples $\langle \mathcal{A}, \varsigma \rangle$, with $\mathcal{A}$ a set of *attributes* $\nu_i$ and $\varsigma$ the *class* of the example, $LearnTree(\text{Tree}(F), \mathcal{A}, \varsigma)$ builds a decision tree $\text{Tree}(F)$ representing a factored representation of the probability $F(\varsigma|\mathcal{A})$.

As shown in Figure 3 (step 3), the reward learning

---

Input: $\text{Tree}(M_t), s, a, s, r$ Output: $\text{Tree}(M_{t+1})$

1. $\text{Tree}(M_{t+1}) \leftarrow \text{Tree}(M_t)$
2. $\mathcal{A} \leftarrow \{x_1, \ldots, x_n\}$
3. $\text{Tree}(R \in M_{t+1}) \leftarrow$ $LearnTree(\text{Tree}(R \in M_t), \mathcal{A} \bigcup \{a\}, r)$
4. For all $i \in |X|$: $\text{Tree}(\text{CPD}^a_{X_i} \in M_{t+1}) \leftarrow$ $LearnTree(\text{Tree}(\text{CPD}^a_{X_i} \in M_t), \mathcal{A}, x'_i)$
5. Return $\text{Tree}(M_{t+1})$

---

Figure 3: SPITI (2): the *Learn*$(s,a,s,r)$ algorithm.

algorithm is straightforward. From an observation of the agent $\langle s, a, r \rangle$ with $s = (x_1, \ldots, x_n)$, we use *LearnTree* to learn a factored representation $\text{Tree}(R)$ of the reward function $R(s,a)$ from the example $\langle \mathcal{A} = (x_1, \ldots, x_n, a), \varsigma = r \rangle$.

The transition model $T$ is composed of a DBN $G_a$ for each action $a$. $G_a$ is quantified with the set of local structures in the conditional probability distributions $\text{CPD}^a = (\text{CPD}^a_{X_1}, \ldots, \text{CPD}^a_{X_n})$[1]. Assuming no synchronic arc in $G_a$ (we have $X'_i \perp\!\!\!\perp X'_j \mid X_1, \ldots, X_n$), SPITI uses *LearnTree* to learn separately a decision tree representation of each $\text{CPD}^a_{X_i}$ from the observation of the agent $\langle s, a, s' \rangle$ with $s = (x_1, \ldots, x_n)$ and $s' = (x'_1, \ldots, x'_n)$, as shown in Figure 3 (step 4). Thus, the explicit representation of the global structure of DBNs representing the transition functions is not built.

The *LearnTree* algorithm has been implemented using ITI (Utgoff et al., 1997) with $\chi^2$ as an information-theoric metric, as described in the next section. We refer to Utgoff et al. (1997) for a detailed description of ITI. $G_a$ is quantified by $\text{Tree}(\text{CPD}^a_{X_i})$ associated to each node $X'_i$. *LearnTree* computes an approximation of the conditional probability $\text{CPD}^a_{X_i}(X'_i|Parents_a(X'_i))$ from the training examples present at each leaf of $\text{Tree}(\text{CPD}^a_{X_i})$ built by ITI. The model $\text{Tree}(M)$ learned is then used in planning (Figure 2) to compute set $S_Q$ of action-value functions.

### 4.3 USING CHI-SQUARE TO DETECT THE DEPENDENCIES

SPITI uses $\chi^2$ as an information-theoric metric to determine the best test $T_{\nu_i}$ to install at a decision node. Once $T_{\nu_i}$ has been selected, we use $\chi^2$ as a test of independence between two probability distributions to avoid training data overfitting (Quinlan, 1986). Thus, a test $T_{\nu_i}$ on the binary attribute $\nu_i$ is installed only if the $\chi^2$ value computed for both probabilities $F(\varsigma|\nu_i = true)$ and $F(\varsigma|\nu_i = false)$ is above a threshold, noted $\tau_{\chi^2}$, determining whether or not the node must be split into two different leaves.

[1] Instead of having a different $\text{Tree}(\text{CPD}^a_{X_i})$ for each action and for each variable, one may maintain only one $\text{Tree}(\text{CPD}_{X_i})$ for each variable by adding the action $a$ to the set of attributes $\mathcal{A}$. We did not consider this case in this paper.

Neither planning nor acting in SPITI require to build an explicit representation of the global structure of DBNs $G_a$. However, as shown in Figure 4, one may build such a representation by assigning to $Parents_a(X'_i)$ the set of variables $X_i$ corresponding to the attributes $\nu_j$ used in the tests in each Tree($\text{CPD}^a_{X_i}$).

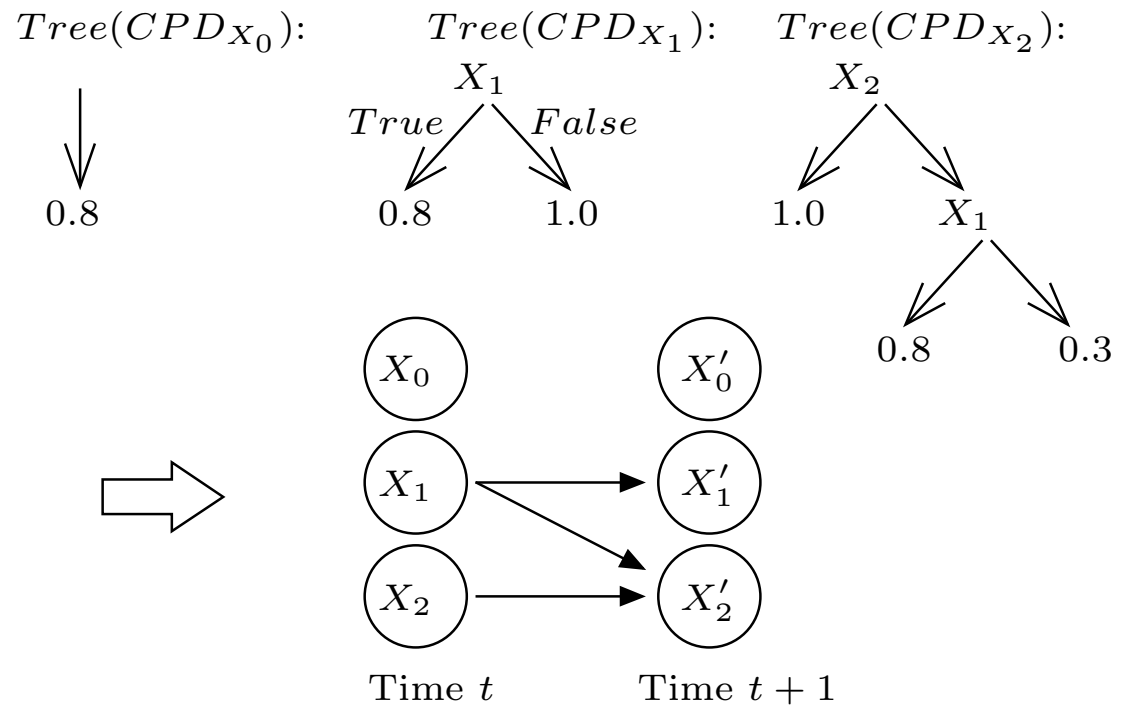

Figure 4: Structure of a DBN $G$ from a set of decision trees {Tree($\text{CPD}_{X_i}$)}. In Tree($\text{CPD}_{X_2}$), the leaf labeled 0.3 means that the probability for $X'_2$ to be true is $P(X'_2|X_1 = False, X_2 = False) = 0.3$.

SPITI is initialized with a set of empty Tree($\text{CPD}^a_{X_i}$), assuming when it starts that the variables $X'_i$ are all independent. When an attribute $\nu_j$ is installed at a decision node, a new dependency of $X'_i$ to the variable $X_j$ associated with $\nu_j$ is defined.

## 4.4 EVALUATING SPITI

We show in Section 5 that SPITI performs better than a stochastic version of DYNA-Q (Sutton, 1990) in terms of discounted reward and size of the model built in structured RL problems. The reward and transition functions and the optimal value function are known for these problems. Based on that knowledge, additional criteria may be used, namely the relative error (Section 4.4.1) and the accuracy of the model (Section 4.4.2) to improve the evaluation of SPITI. These criteria respectively measure how good a policy is compared to an optimal policy, and how accurate is the model of transitions learned by SPITI.

### 4.4.1 Relative Error

The optimal value function $V^*$, computed off-line using SVI, may be used as a reference to evaluate a policy. We define the relative error, noted $\xi_\pi$, between $V^*$ and $V_\pi$ as the average of the relative value error $\Delta V = (V^* - V_\pi)/V^*$ (with $V^* \geq V_\pi$). We compute $\xi_\pi$ with operators using tree representations. Given a policy Tree($\pi$) to evaluate, we use the *Structured Successive Approximation* (SSA) algorithm (Boutilier et al., 2000) based on the exact transitions and reward functions to compute its associated value function Tree($V_\pi$). Then, from set $S_{\Delta V} = \{\text{Tree}(V^*), \text{Tree}(V_\pi)\}$, we first compute Tree($\Delta V_\pi$) using the $Merge(S_{\Delta V})$ operator and using as combination function $\Delta V$, the relative value error. Then, $\xi_\pi$ is computed according to:

$$\xi_\pi = \frac{\sum_{l \in \text{Tree}(\Delta V_\pi)} \Delta V_l \cdot S_l}{\prod_{i \in |X|} |Dom(X_i)|} \tag{2}$$

with $\Delta V_l$ the label of the leaf $l$ and $S_l$ the size of the state space represented by $l$.

### 4.4.2 Accuracy of the Model

We introduce the measure $\mathcal{Q}_{\chi^2}$ to qualify the accuracy of the model learned by SPITI. The accuracy of the model is complementary to the relative error because it evaluates the model learned in SPITI independently of the reward function. $\mathcal{Q}_{\chi^2}$ is defined as:

$$\mathcal{Q}_{\chi^2} = \frac{\sum_{a \in A} \sum_{i \in |X|} \sigma_{a,i}}{|A| * \prod_{i \in |X|} |Dom(X_i)|} \tag{3}$$

with $\sigma_{a,i}$ defined in Figure 5.

---

Input: $a, i \in |X|$ Output: $\sigma_{a,i}$

1. $\sigma_{a,i} = 0$
2. $Merge(\{Tree_{def}(\text{CPD}^a_{X_i}), Tree_t(\text{CPD}^a_{X_i})\})$ using as a combination function:

   $\sigma_{a,i} = \sigma_{a,i} + Q(\chi^2_{(X'_i,a,s')}) \cdot S_l$

   with $S_l$ the size of the state space represented by $l_t$ and $Q(\chi^2_{(X'_i,a,s')})$ the probability associated with the value $\chi^2_{(X'_i,a,s')}$ computed from the probability $P_{def}(X_i)$ labeling the leaf in $Tree_{def}(\text{CPD}^a_{X_i})$ and $P_t(X'_i|s,a)$ labeling the leaf $l_t$
3. Return $\sigma_{a,i}$

---

Figure 5: Computation of $\sigma_{a,i}$ used in the evaluation of the model learned by SPITI.

The values $\chi^2_{(X'_i,a,s')}$ and $Q(\chi^2_{(X'_i,a,s')})$ are computed using implementations proposed in Press et al. (1992) with 1 degree of freedom. The probability $Q(\chi^2_{(X'_i,a,s')})$ is computed for each leaf and weighted with the size of the state space represented by the leaf, penalizing errors in the model that covers a large part of the state space. The average is obtained by dividing the weighted sum by the number of state/action pairs.

# 5 RESULTS

We present three different analyses based on empirical evaluations of SPITI. The first analysis studies the relation between the value of the threshold $\tau_{\chi^2}$ and the size of the model built by SPITI on one hand, and between $\tau_{\chi^2}$ and the relative error of the value function of the induced greedy policy on the other hand. The second analysis compares SPITI to DYNA-Q in terms of discounted reward and model size. The last analysis studies the generalization property of the model learning process in SPITI.

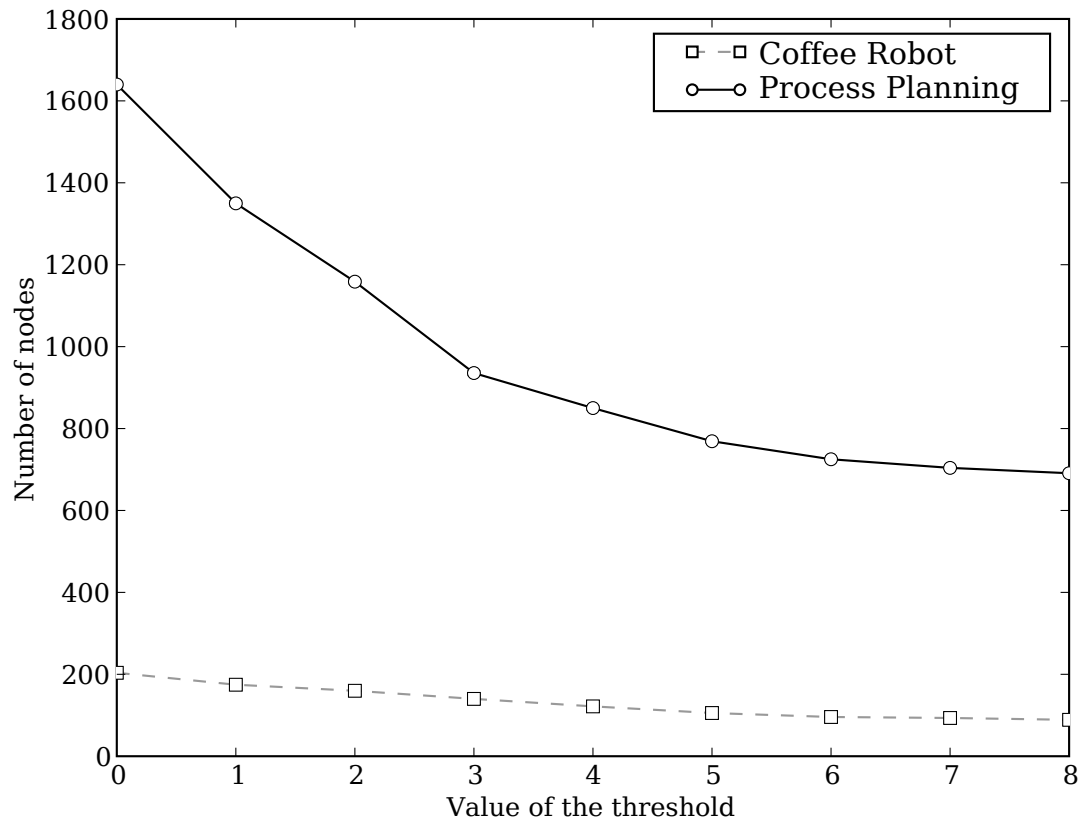

Figure 6: Number of nodes of the model learned in SPITI on the *Coffee Robot* and *Process Planning* problems. A high value of $\tau_{\chi^2}$ decreases significantly the size of the model learned.

These analyses are run on a set of stochastic problems defined in Boutilier et al. (2000). A set of initial states and a set of terminal states are added to the problem definitions to let the agent perform multiple trials during an experiment. When an agent is in a terminal state, its new state is randomly initialized in one of the initial states. The set of initial states is composed of all the non-terminal states from which there is at least one policy reaching a terminal state. We run 20 experiments for each analysis. When required, we use SVI to compute off-line optimal policy using the span semi-norm as a termination criterion and SSA with the supremum norm as a termination criterion. We use $\gamma = 0.9$ for both algorithms.

## 5.1 CHI-SQUARE THRESHOLD

In order to study the influence of $\tau_{\chi^2}$ on the quality of the policy, we use the following protocol: first, a random trajectory $\mathcal{J}$ is executed for $T = 4000$ time steps. Second, the value of $\tau_{\chi^2}$ is fixed and a transition and reward model $M_{\mathcal{J},\tau}$ is built from the trajectory by SPITI as described in Section 4.2. Third, a policy $\pi_{\mathcal{J},\tau}$ based on $M_{\mathcal{J},\tau}$ is computed off-line using SVI. Finally, we compute the relative error $\xi_\pi$ as described in Section 4.4.1.

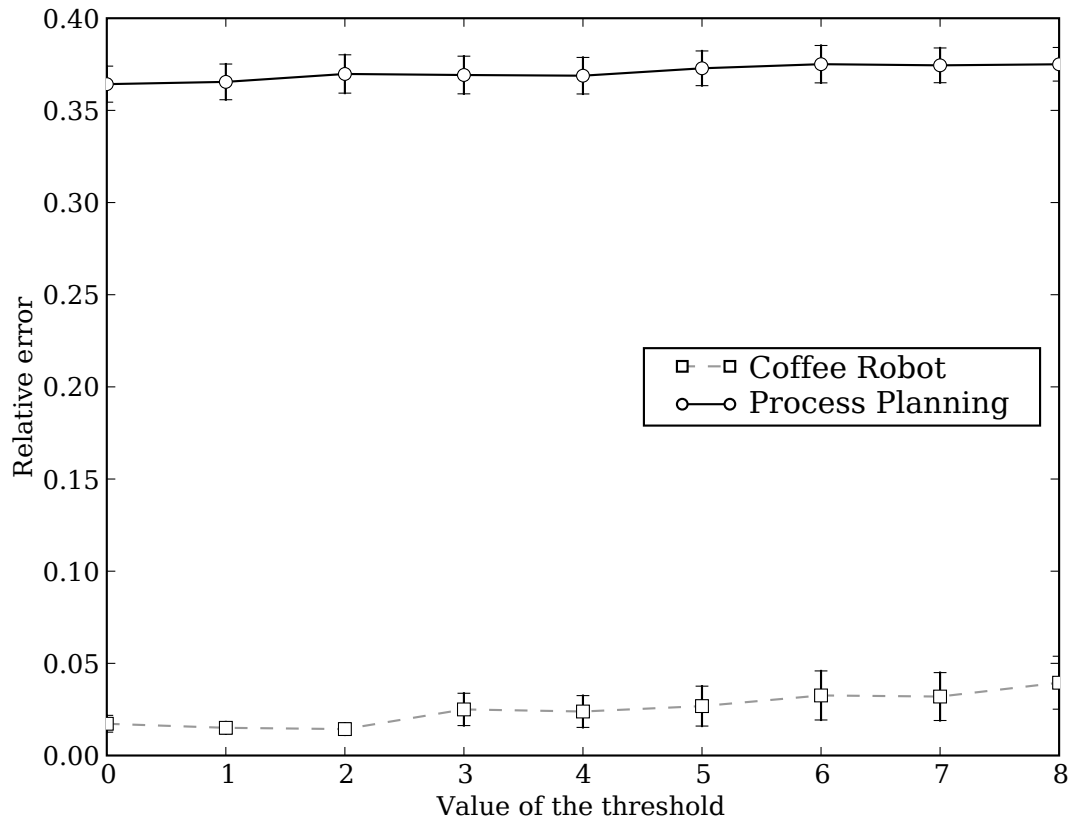

Figure 7: Relative error induced from the model learned in SPITI on the *Coffee Robot* and *Process Planning* problems. The value of $\tau_{\chi^2}$ has a limited impact on the policy generated by SPITI.

The first empirical study is done on *Coffee Robot* and *Process Planning*. Both problems are defined in Boutilier et al. (2000) and the complete definition of their reward and transition functions is available on the SPUDD website[2]. *Coffee Robot* is a stochastic problem composed of 4 actions and 6 boolean variables. It represents a robot that must go to a café and buy some coffee to deliver it to its owner. The robot reaches a terminal state when its owner has a coffee. *Process Planning* is a stochastic problem composed of 14 actions and 17 binary variables ($1,835,008$ state/action pairs). A factory must achieve manufactured components by producing, depending on the demand, high quality components (using actions such as hand-paint or drill) or low quality components (using actions such as spray-paint or glue).

Figure 6 shows the size of the transition model built by SPITI on the *Coffee Robot* and *Process Planning* problems for different values of $\tau_{\chi^2}$. It illustrates the impact of $\tau_{\chi^2}$ on the number of nodes created in the trees and, consequently, on the number of dependencies between the variables of the FMDP. On both problems, the size of the model is at least divided by 2 for high values of $\tau_{\chi^2}$ as compared to low values.

Whereas the value of $\tau_{\chi^2}$ has a significant impact on the size of the model, it has a much more limited impact on the generated policy $\pi_{\mathcal{J},\tau}$, as shown in Figure 7. Despite decreasing model sizes, the relative error $E_{\mathcal{J},\tau}$ computed for $\pi_{\mathcal{J},\tau}$ increases only slightly on both *Coffee Robot* and *Process Planning* problems.

[2] http://www.cs.ubc.ca/spider/jhoey/spudd/spudd.html

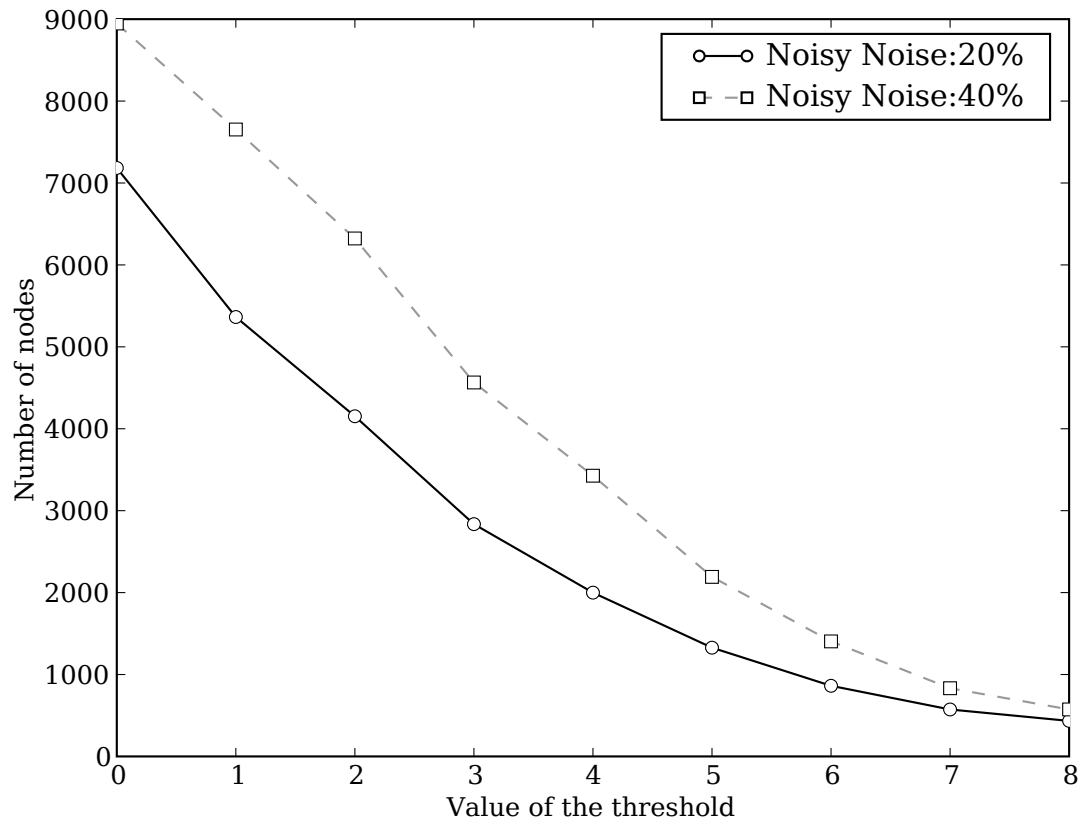

Figure 8: Size of the model learned in SPITI on the *Noisy* problems with two different levels of noise. A high value of $\tau_{\chi^2}$ decreases significantly the size of the model learned.

To examine the consequences of the threshold $\tau_{\chi^2}$ on a very stochastic problem, we define a problem named *Noisy*. Boutilier et al. (2000) define two problems, namely *Linear* and *Expon*, to illustrate respectively the best and worst case scenario for SPI. The transition and reward functions of *Noisy* are defined according to the definition of the *Linear* problem with a constant level of noise on all existing transitions. We present additional results about SPITI in the *Linear*, *Expon* and *Noisy* problems in Section 5.3.

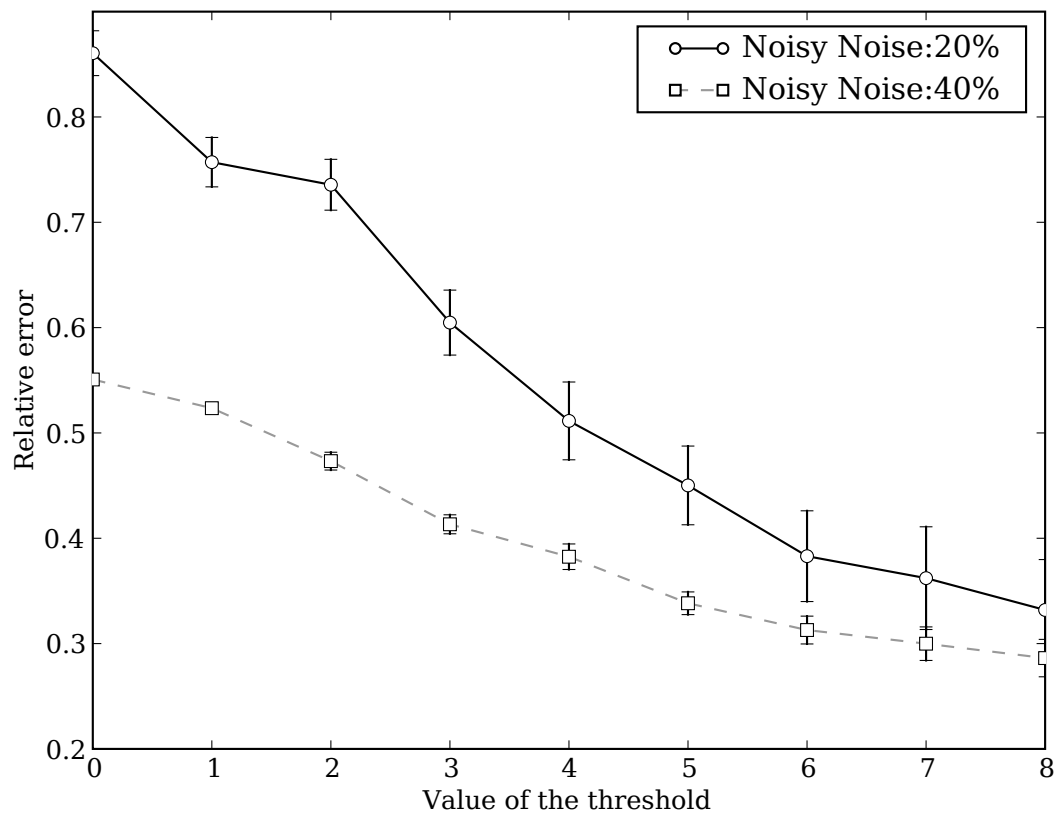

Figure 9: Relative error induced from the model learned in SPITI on the *Noisy* problem with two different levels of noise.

Figure 8 shows that for a very stochastic problem, namely *Noisy* with two levels of noise, 20% and 40%, with a fixed size of 8 binary variables and actions, the impact of the threshold $\tau_{\chi^2}$ is more important than in the previous problems *Coffee Robot* and *Process Planning* which contain some deterministic transitions. SPITI builds a model from more than 9000 nodes to less than 300 nodes for an identical trajectory. Figure 9 shows that on the *Noisy* problem, a more compact transition model generates a better policy than a larger transition model, even if this model has been learned from the same trajectory.

## 5.2 DISCOUNTED REWARD

In this study, we compare SPITI to a stochastic implementation of DYNA-Q (Sutton, 1990) on *Coffee Robot* and *Process Planning*. We use $\gamma = 0.9$ and a $\epsilon$-*greedy* exploration policy with a fixed $\epsilon = 0.1$ for both DYNA-Q and SPITI. In DYNA-Q, we used $\alpha = 1.0$, the number of planning steps is set to twice the size of the model, and $Q(s, a)$ is initialized optimistically (with high values). In SPITI, the results of the previous section show that a high value for the threshold $\tau_{\chi^2}$ is appropriate. Thus, we set $\tau_{\chi^2} = 7,88$ (corresponding to a probability of independence of 0.995).

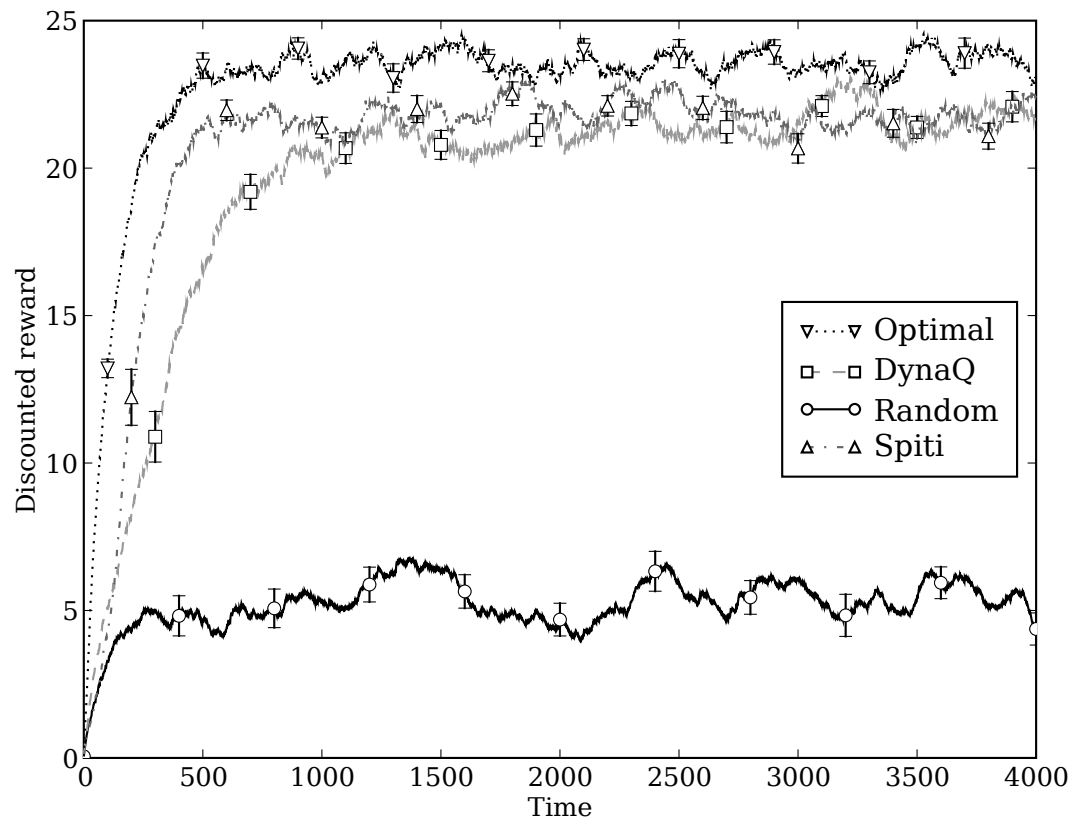

Figure 10: Discounted reward obtained on the *Coffee Robot* problem. DYNA-Q and SPITI execute quickly a near optimal policy on this small problem.

We also use as reference two agents noted RANDOM and OPTIMAL, executing respectively a random action and the best action at each time step. The discounted reward is defined as $R_t^{disc} = r_t + \gamma' R_{t-1}^{disc}$ with $r_t$ the reward received by the agent and $\gamma' = 0.99$[3].

Figure 10 shows the discounted reward $R^{disc}$ obtained by the agents on the *Coffee Robot* problem. On this small problem, both DYNA-Q and SPITI quickly execute a near optimal policy. However, the model learned by

[3] We use $\gamma \neq \gamma'$ to make the results illustrating $R^{disc}$ more readable.

SPITI is composed of approximately 90 nodes whereas DYNA-Q builds a model of 128 nodes, that is the number of transitions in the problem[4].

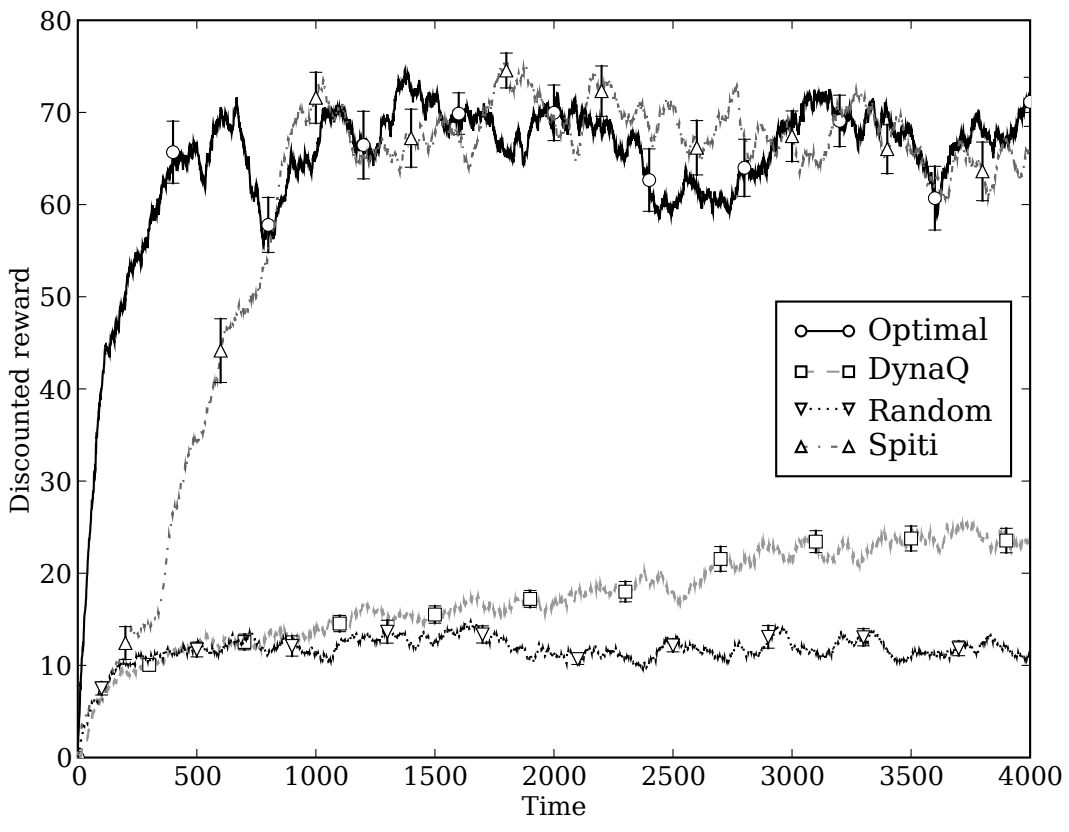

Figure 11: Discounted reward obtained on the *Process Planning* problem. SPITI executes quickly a near optimal policy in a large structured problem, unlike DYNA-Q.

Figure 11 shows the discounted reward $R^{disc}$ obtained by the agents on the *Process Planning* problem. SPITI is able to execute a near optimal policy in approximately 2000 time steps, whereas DYNA-Q only starts to improve its policy after 4000 time steps. Comparing the size of the transition model learned, DYNA-Q builds a representation of approximately 2500 nodes which would keep growing if the experiment was continued whereas SPITI builds a structured representation stabilized to less than 700 nodes.

### 5.3 GENERALIZATION IN MODEL LEARNING IN SPITI

In this third study, we use $\mathcal{Q}_{\chi^2}$ to qualify the loss of accuracy of the model built by the *Learn* algorithm (Figure 3) when the size of a problem grows whereas the experience of the agent does not. We use the following protocol: first, a random trajectory $\mathcal{J}$ is executed in the environment for $T = 4000$ time steps. Then, we compute $\mathcal{Q}^{\mathcal{J}}_{\chi^2}$ with the transition model $M^T_{\mathcal{J}}$ built from the trajectory $\mathcal{J}$ by SPITI (as described in Section 4.2) and the actual definition of the problem. Finally, we restart using the same problem with one more action and one more binary variable. We use a random trajectory in this experiment to avoid any dependency to the reward function learning process.

The experiment is run on the three following problems: *Linear*, *Expon* and *Noisy*, using a level of noise of 20%. We used $\tau_{\chi^2} = 7,88$ to build the model $M^T_{\mathcal{J}}$. The size of the problem grows from $2^4 \cdot 4 = 64$ to $2^{20} \cdot 20 = 20,971,520$ state/action pairs. Figure 12 shows that the accuracy $\mathcal{Q}_{\chi^2}$ of the model built by SPITI decreases linearly with the size of the problems.

[4]A terminal state does not have transitions.

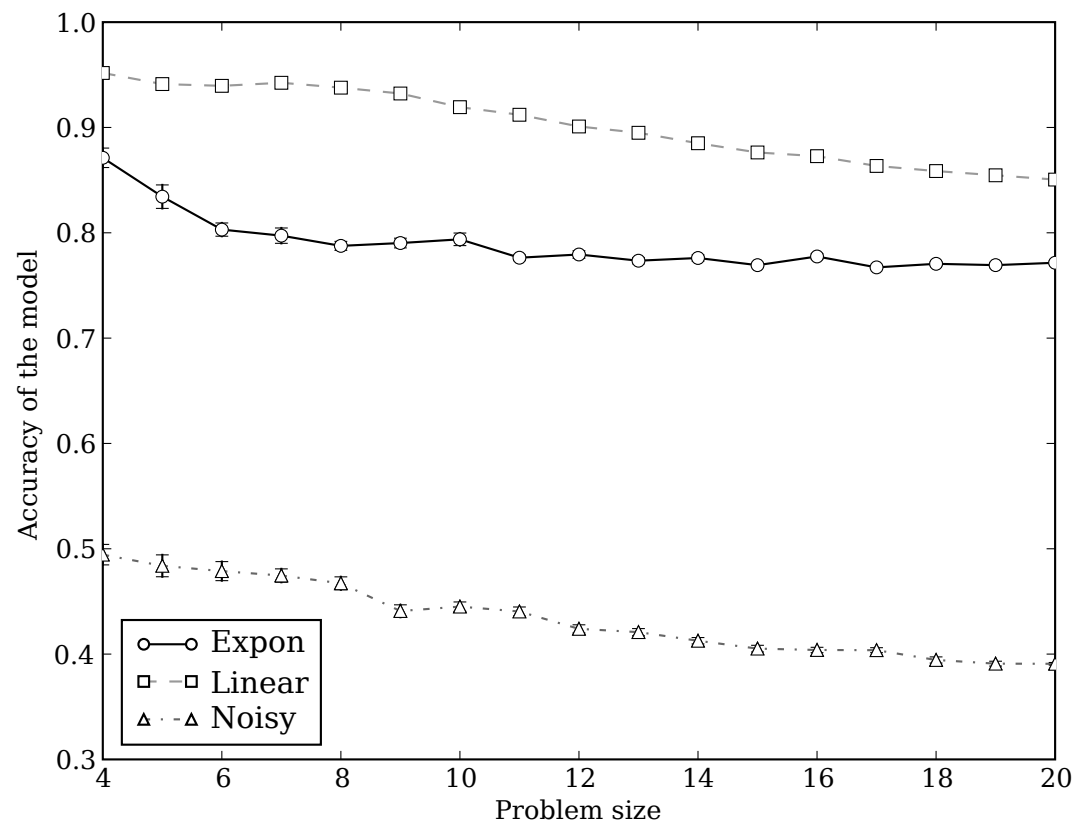

Figure 12: Accuracy of the model learned by SPITI after $T = 4000$ time steps on problems with variable size. The accuracy decreases linearly whereas the size of the problem grows exponentially (from 64 to 20,971,520 state/action pairs).

## 6 DISCUSSION

SDYNA is an architecture designed to integrate planning, learning and acting in FMDPs. In this paper, we have focused on SPITI, an instantiation of SDYNA, that uses decision trees as factored representation. SPITI simultaneously learns a structured model of the reward and transition functions and uses an incremental version of SVI to compute its policy.

SPITI learns the structure of a RL problem using $\chi^2$ as a test of independence between two probability distributions. We have first shown that the threshold $\tau_{\chi^2}$ used to determine whether or not new dependencies should be added to the transition model has a significant impact on the size of the model learned and a more limited impact on the quality of the policy generated by SPITI (Section 5.1). Moreover, figure 8 and figure 9 show that setting $\tau_{\chi^2}$ to a low value may not be adapted in very stochastic problems. Indeed, large models may contain unnecessary dependencies and, consequently, require more samples to be accurately quantified. When setting $\tau_{\chi^2}$ to a high value, SPITI is able to build a compact representation of the transition and reward functions of the problem without degrading its policy.

Second, we have shown that SPITI is able to learn the structure of RL problems with more than one million state/action pairs and performs better than DYNA-Q. Unlike tabular learning algorithms, decision tree induction algorithms build factored representations that endow the agent with a generalization property. The decision trees used to represent the transition and reward functions in SPITI propose a default class distribution for examples that have not been presented. Consequently, an agent may be able to choose adequate actions in states not visited yet. As we have shown in Section 5.2, the generalization property in SPITI accelerates the resolution of large RL problems.

Third, we have used an accuracy measure to study the generalization property of transition model learning in SPITI. We have shown that for a constant number of time steps, the accuracy of the model built by SPITI decreases linearly when the number of state/action pairs in the problem grows exponentially. This result indicates that SPITI is able to scale well in larger structured RL problems.

SPITI has been evaluated using three different criteria: the relative error $\xi_\pi$, the discounted reward $R^{disc}$ and the accuracy measure $\mathcal{Q}_{\chi^2}$. Two of these criteria cannot be applied in real world RL problems: the relative error requires to know the optimal value function whereas the accuracy measure requires to know the exact transition function of the problem. Moreover, the discounted reward is the only criterion that takes into account the loss of reward received due to the exploration policy. Thus, $\xi_\pi$ and $\mathcal{Q}_{\chi^2}$ are complementary to $R^{disc}$ that fully evaluates all the parameters of SPITI and may be used on all kind of RL problems.

# 7 CONCLUSION AND FUTURE WORK

We have described in this paper a general approach to model-based RL in the context of FMDPs, assuming that the structure of problems is unknown. We have presented an instantiation of this approach called SPITI. Our results show empirically that SPITI performs better than a tabular model-based RL algorithm by learning a compact representation of the problem from which it can derive a good policy, exploiting the generalization property of its learning method, particularly when the problem gets larger.

However, SPITI is currently limited by its exploration policy, *$\epsilon$-greedy*, and its planning method, adapted from SVI. We are currently working on integrating other model-based learning (Guestrin et al., 2002) and planning (Hoey et al., 1999; Guestrin et al., 2003) methods in FMDPs to address larger problems than those addressed in this paper.

# ACKNOWLEDGEMENT

We wish to thank Christophe Marsala and Vincent Corruble for useful discussions. Thanks to the anonymous referees for their helpful suggestions.